\documentclass[10pt,twocolumn,letterpaper]{article}

\usepackage{cvpr}
\usepackage{times}
\usepackage{epsfig}
\usepackage{graphicx}
\usepackage{amsmath}
\usepackage{amssymb}
\usepackage{multirow}
\usepackage{graphicx}
\usepackage{booktabs} 
\usepackage{color}
\usepackage{bm}
\usepackage{etoolbox}
\makeatletter
\patchcmd{\maketitle}
 {\def\@makefnmark}
 {\def\@makefnmark{}\def\useless@macro}
 {}{}
\makeatother

\usepackage[pagebackref,breaklinks,colorlinks]{hyperref}

\cvprfinalcopy % *** Uncomment this line for the final submission

\def\cvprPaperID{****} % *** Enter the CVPR Paper ID here

\begin{document}

 \author{Naiyuan Liu$^{1,2}$, Xiaohan Wang$^{1}$, Xiaobo Li$^{3}$, Yi Yang$^{1}$, Yueting Zhuang$^{1}$ \\
 \texttt{\small naiyuan.liu@student.uts.edu.au,xiaohan.wang@zju.edu.cn, yangyics@zju.edu.cn} \\
$^1$ReLER Lab, CCAI, Zhejiang University, $^2$University of Technology Sydney,  $^3$Alibaba Group
 }

\title{ReLER@ZJU-Alibaba Submission to the Ego4D Natural Language Queries Challenge 2022}

\maketitle
%\thispagestyle{empty}

%%%%%%%%% ABSTRACT
\begin{abstract}
In this report, we present the ReLER@ZJU-Alibaba submission to the Ego4D Natural Language Queries (NLQ) Challenge in CVPR 2022. Given a video clip and a text query, the goal of this challenge is to locate a temporal moment of the video clip where the answer to the query can be obtained. 
To tackle this task, we propose a multi-scale cross-modal transformer and a video frame-level contrastive loss to fully uncover the correlation between language queries and video clips.
Besides, we propose two data augmentation strategies to increase the diversity of training samples. The experimental results demonstrate the effectiveness of our method. The final submission ranked first on the leaderboard. The code is available at {\url{https://github.com/NNNNAI/Ego4d_NLQ_2022_1st_Place_Solution}}.

\end{abstract}

\section{Introduction}
Given a video clip and a text query, the goal of Ego4D NLQ task \cite{grauman2021ego4d} is to locate the corresponding moment span where the answer to the query can be obtained.
There are two challenges to Ego4D NLQ task: extremely long duration time of videos and shortage of videos. 
First, the total duration of the video clips on Ego4D is extremely long while the duration of moments span represents a very small percentage of the total duration. 
For example, the average duration of the video clip is up to 7.5 minutes while the average duration of the span is less than 5 seconds. 
The second issue is shortage of videos.
Concretely, Ego4D NLQ training dataset has more than 10000 video clip-text pairs, but there are only about 1200 union video clips which are not enough to learn such a complex task as NLQ.

To alleviate both challenges, we propose a multi-scale cross-modal transformer making the video features interact with text features more adequately.
In addition, video frame-level contrastive loss is introduced to enforce our model to focus on video frames that fall into moment span. 
To solve the challenge of video shortage, we  propose two data augmentation methods: variable-length sliding window sampling (SW) and video splicing (VS), to collect more samples and avoid overfitting issue. 
Our method outperforms previous state-of-the-art methods and achieves the best performance on the test sets.

%------------------------------------------------------------------------
\section{Related Work}

The NLQ task can be treated as a multi-modal retrieval task. There exist similar tasks, including moment retrieval~\cite{zhang2021multi,zhang2020span}, video highlight detection~\cite{lei2021qvhighlights} and text-video retrieval~\cite{wang2021t2vlad,zhao2022centerclip}. These tasks retrieve or localize a temporal moment that semantically corresponds to a given language query. 
Many previous works \cite{zhang2021multi,zhang2020learning,zhang2021natural,zhang2020span,lei2021qvhighlights} aim to enhance the interactions among multiple knowledge representations from different modalities \cite{yang2021multiple} on these tasks. However, the NLQ task is more challenge due to the long duration of videos and video shortage. Moreover, a good video representation \cite{girdhar2022omnivore,feichtenhofer2019slowfast} can make the NLQ task easier.
For example, CLIP \cite{radford2021learning} demonstrates the benefits of large-scale text-image pre-training. Our idea is to exploit a more efficient way for long videos.

\begin{figure*}[t]
\vspace{-4mm}
\center
\includegraphics[width=1\linewidth]{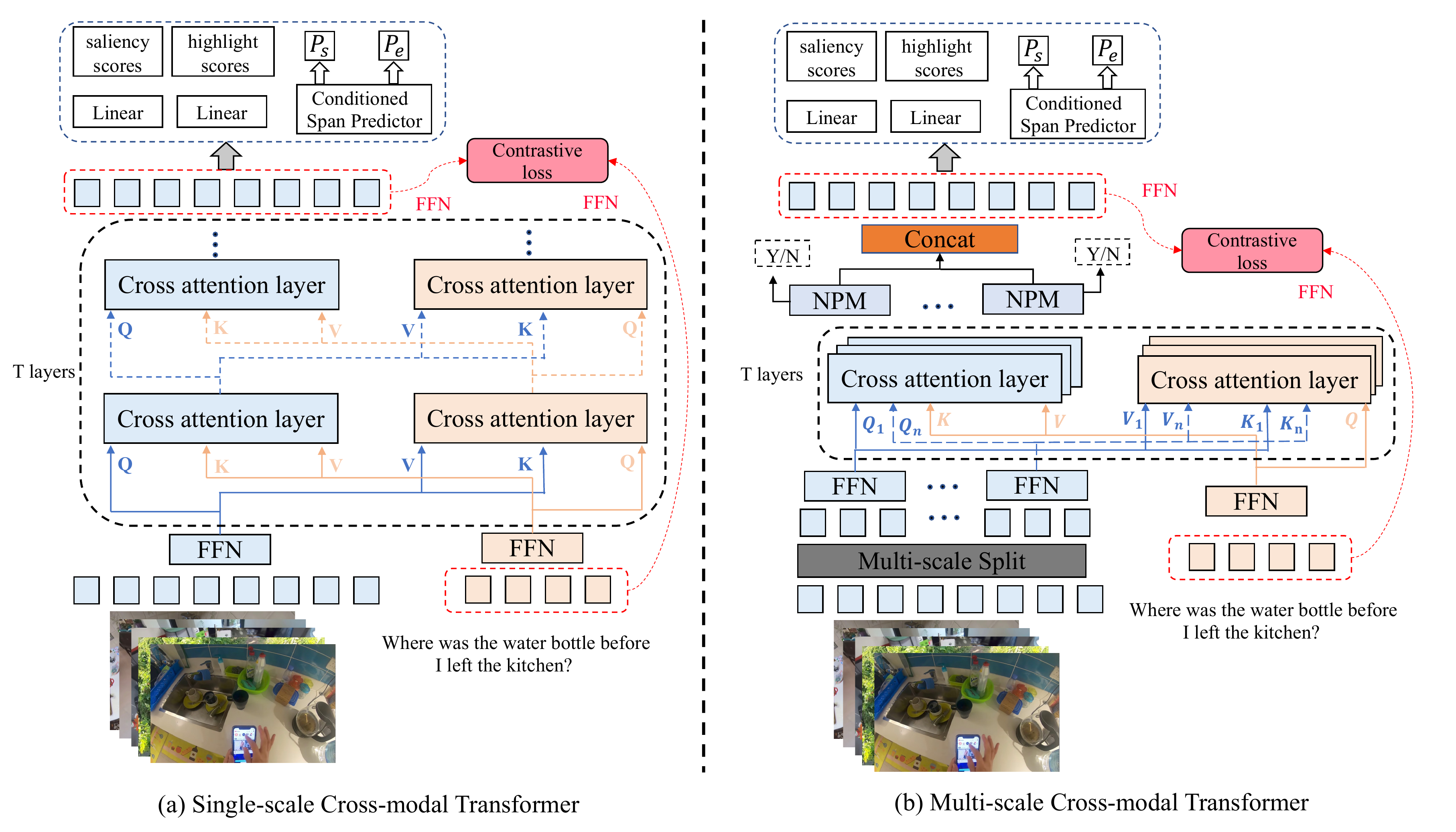}
\caption{
The overall framework of our approach. (a) depicts our single-scale cross-modal transformer. (b) shows the details of our multi-scale cross-modal transformer.
}
%\vspace{-2mm}
\label{fig:framework}  
\end{figure*}
% %-------------------------------------------------------------------------

\section{Methodology}

As shown in Figure ~\ref{fig:framework}, we use $T$ cross-attention layers to build our multi-scale cross-modal transformer as the backbone.  
Then, we build a saliency scores predictor \cite{lei2021qvhighlights}, a highlight region predictor \cite{lei2021qvhighlights}, and a conditioned span predictor \cite{zhang2020span} upon the backbone. 
We use pre-extracted features for both video and text input.
Additionally, we utilize pre-extracted video and text features as inputs of the backbone.
To avoid overfitting issue, two data augmentation methods are adopted during training, including video splicing (VS) and variable-length sliding window sampling (SW). 

\subsection{Input preparation}
We use Slowfast features \cite{feichtenhofer2019slowfast} and Omnivore features \cite{girdhar2022omnivore} provided by Ego4D developers as video features. Sepecifically, Slowfast uses window size 32 and temporary stride 16 to extract features (roughly two frames of Slowfast features per second for 30-fps videos). In addition, Omnivore uses window size 32 and temporary stride 6 to extract features (roughly five frames of Omnivore features per second for 30-fps videos). 
We introduce CLIP \cite{radford2021learning} feature to improve cross-modal representation learning.
For each frame of video features, we randomly select one of its input frames and then feed it to the image encoder (ViT-B/16) of CLIP to get the CLIP visual feature. 
The video features and CLIP visual features are concatenated along channel dimension as the final video input. 
A text input is obtained by the CLIP text encoder. 
Instead of taking the EOS token as an aggregate representation of text, we reserve text sequence length to use text token-level information.

\begin{figure}[t]
\center
\includegraphics[width=\columnwidth]{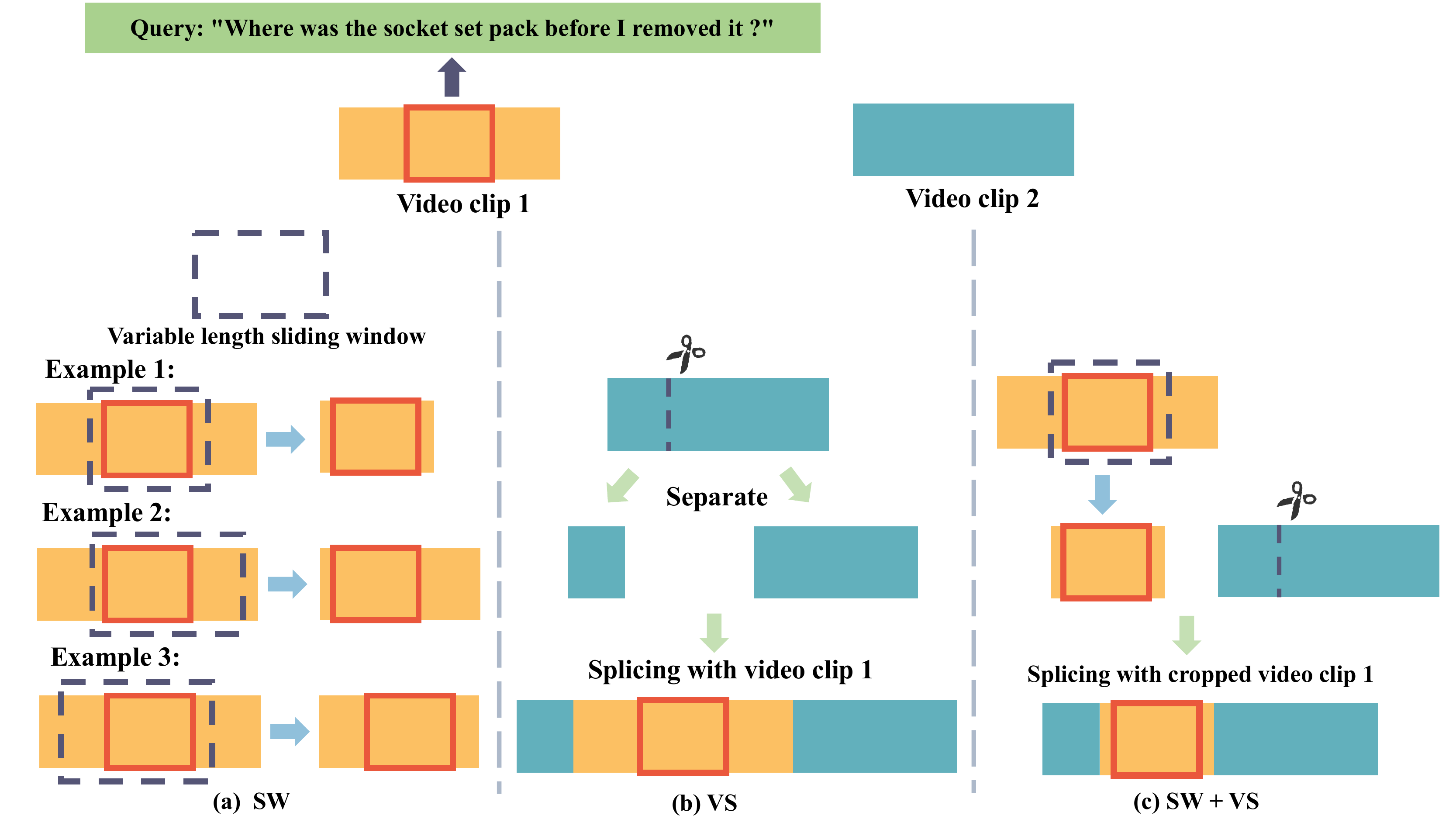}
\caption{
Illustration of data augmentation. (a) shows how the variable-length sliding window sampling strategy (SW) works. (b) shows how the video splicing strategy (VS) works. (c) is a combination of these two data augmentations which leads to better performance.
}
\vspace{-2mm}
\label{fig:data_argu}  
\end{figure}

\subsection{Multi-scale Cross-modal Transformer}
\textbf{Cross-attention mechanism.} 
Single-scale cross-modal transformer and multi-scale cross-modal transformer are both built by a stack of $T$ cross-attention layers shown in Figure ~\ref{fig:framework}. 
We set $T=3$ by default. 
The structure of the cross-attention layer is the same as that of the standard transformer encoder block ~\cite{vaswani2017attention}, including a multi-head attention layer and a position-wise fully connected feed-forward network. 
To fully uncover the correlation between video and text, we use a cross-attention mechanism instead of the self-attention mechanism which is used by the standard transformer encoder block.
Query, key, and value are all obtained from the same input modality through three linear layers in self-attention mechanism. 
However, the cross-attention mechanism exchanges the key-value pairs of different input modalities for attention operation as shown in Figure ~\ref{fig:framework} (a). 
As a result, we get the attention that has been language-conditioned in the video stream and attention that has been video-conditioned in the linguistic stream. 
This mechanism ensures that each video feature interacts with text features independently, regardless of the length of the video. We only feed the final feature in the video stream to prediction heads.

\textbf{Multi-scale mechanism.} We build our multi-scale cross-modal transformer by adopting multi-scale split-and-concat strategy from VSLNet-L \cite{zhang2021natural} as shown in Figure ~\ref{fig:framework} (b). 
Here, we summarize the key idea of this strategy below.
This strategy splits a video into K video segments: $\bm{V}=[\bm{V}_{1},\dots,\bm{V}_{K}]$. 
Each video segment $\bm{V}_{k}$ is fed to the cross-modal transformer separately and produces feature $\bm{F}_{k}$. Each feature $\bm{F}_{k}$ is then processed by Nil Prediction Module (NPM) \cite{zhang2021natural} and produces a score $\mathcal{S}_{\text{nil}}^{k}$, which indicates the confidence of video segment $\bm{V}_{k}$ overlaps with query corresponding moment span. All features $\bm{F}_{k}$ are re-weighted by $\mathcal{S}_{\text{nil}}^{k}$ and produce $\bm{\bar{F}}_{k}$.
All $\bm{\bar{F}}_{k}$ are concatenated into $\bm{\bar{F}}_{final}$ along the sequence dimension. In the end, we send $\bm{\bar{F}}_{final}$ to prediction heads. Please refer to VSLNet-L \cite{zhang2021natural} for details about this Strategy.

To estimate the target moment span, we use the conditioned span predictor and highlight predictor from VSLNet \cite{zhang2020span}, and we also use the saliency predictor following Moment DETR \cite{lei2021qvhighlights}. Sepecifically, the conditioned span predictor is constructed with two transformer encoder layers and two linear layers to predict the start and end boundary of the moment span. The saliency predictor and highlight predictor are both built with two linear layers to predict which video frame feature falls into the moment span.

\subsection{Video Frame-level Contrastive Loss} 
The goal of the Ego4D NLQ task is to locate the moment span by using the text information. 
The similarity between text features and video frame features belonging to the moment span should be higher than the similarity between text features and video frame features that fall out of the moment span. Therefore, we introduce video frame-level contrastive loss.
The similarity calculation function between video frame feature and text embedding is as follows:

\begin{align}\label{eq:rew}
  F(v,T) = \frac{\sum_{\bm{t_j}\in{T}} v \cdot t_j/\tau}{|T|},
\end{align}
where  $v \in \mathbb{R}^{1\times d_v}$ is the single video frame from the whole video clip sequence, and $T \in \mathbb{R}^{L_t\times d_t}$ is the whole text embedding sequence ( $t_j \in \mathbb{R}^{1\times d_t}$), $\tau$ is the temperature hyper-parameter. We set $\tau=0.07$ by default. 
\begin{equation}\small\label{eq:pNCE}
\!\!\!\!\mathcal{L}^{\text{NCE}}_i\!=\!\frac{1}{|\mathcal{P}_i|}\!\!\sum_{\bm{v_i}^+\in\mathcal{P}_i\!\!}\!\!\!-_{\!}\log\frac{\exp(F(\bm{v_i}^{+\!\!},\bm{T_i}))}{\exp(F(\bm{v_i}^{+\!\!},\bm{T_i}))
+\!\sum\nolimits_{\bm{v_i}^{-\!}\in\mathcal{N}_i\!}\!\exp(F(\bm{v_i}^{-\!\!},\bm{T_i}))},\!\!
\vspace{-2pt}
\end{equation}
where $\mathcal{P}_i$ and $\mathcal{N}_i$ denote video embedding collections of the positive and negative frames of $i^{th}$ video-text pair. A video frame feature is a positive sample if it falls into the moment span. In contrast, a video frame feature that falls out of the moment span is a negative sample. The total contrastive loss is as follows:
\vspace{-1pt}
\begin{equation}\small\label{eq:com}
\mathcal{L}_{\text{NCE}}=\sum\nolimits_i\mathcal{L}^{\text{NCE}}_i ,
\vspace{-1pt}
\end{equation}
The total loss of our method is shown blow:
\begin{equation}
    \mathcal{L} = \mathcal{L}_{\text{span}} + \mathcal{L}_{\text{QGH}} + \mathcal{L}_{\text{NPM}} + \mathcal{L}_{\text{saliency}}+ \mathcal{L}_{\text{NCE}}, 
\end{equation}
% Please go to the corresponding paper \cite{zhang2020span, lei2021qvhighlights} for losses other than NCEloss
For losses other than our contrastive loss, please refer to the corresponding paper \cite{zhang2020span, lei2021qvhighlights, zhang2021natural} for more details.

\subsection{Data augmentation} 
Even though the Ego4D NLQ training dataset has more than 10000 video clip-text pairs, there are only about 1200 union video clips which are not enough to learn such a complex task as NLQ.
In order to get more video clip data to facilitate convergence and avoid overfitting issue, we design a new data augmentation method by inserting positive clips into null video clips. 
Positive clips are sampled from a long video with variable background padding so as to increase diversity. This approach is a combination of two basic methods: variable-length sliding window sampling strategy (SW) and video splicing strategy (VS), as shown in Figure ~\ref{fig:data_argu}.

\textbf{Variable-length sliding window sampling strategy}. 
Inspired by MS 2D-TAN \cite{zhang2021multi}, we propose a variable-length sliding window sampling strategy to get more positive clips during training, as shown in Figure ~\ref{fig:data_argu} (a). 
Specifically, we define a length ratio interval [$r_s$, $r_e$]. Suppose we sample a video $V$ whose length is $l_v$. Then we will randomly sample a ratio $\hat{r}$ from the length ratio interval ($r_s \leq \hat{r} \leq r_e$). The sliding window size is equal to $\hat{r} * l_v$. We use this sliding window to generate positive clip $V_p$ from the video $V$, and we ensure that the generated positive clip contains the whole query corresponding moment span. 

\textbf{Video splicing strategy}. 
Another data augmentation method is to insert one video clip into a null video clip, as shown in Figure ~\ref{fig:data_argu} (b). Specifically, we sample two videos $V_1$ and $V_2$ each time. We randomly select a cut-in position on $V_2$, divide the video into two parts $V_{21}$ and $V_{22}$, and place $V_{21}$ and $V_{22}$ on the head and tail of $V_1$ respectively to generate a new video clip.
There is a hyper-parameter called splicing probability $P_{vs}$ to control whether to splice $V_2$  and $V_1$ together for this sampling. 

\textbf{Combination of these two data augmentation}.
 We combine these two methods as our final data augmentation method, as shown in Figure ~\ref{fig:data_argu} (c). 
In the experiments section \ref{sec:Experiments}, we observe that combining these two methods gains better performance than using any of them alone. Similarly, we first sample two videos $V_1$ and $V_2$. Moreover, we adopt the variable-length sliding window sampling strategy for $V_1$ to obtain the positive clip $V_{1p}$, and then utilize the video slicing strategy for $V_{1p}$ and $V_2$ to achieve the final video clip.
We set the length ratio interval to [0.4,0.8] and set the splicing probability $P_{vs}$ to 0.5. We found this to be the best value for these two hyper-parameters. It is worth noting that these data augmentation strategies are only used during the training stage.

\section{Experiments}
\label{sec:Experiments}
\begin{table}[t]
    \small
    \centering
    \begin{tabular}{c c c c c}
        \toprule
        \multirow{2}{*}{Method} & \multicolumn{2}{c}{IoU=$0.3$ (\%)} & \multicolumn{2}{c}{IoU=$0.5$ (\%)}\\
        & R@1 & R@5 & R@1 & R@5\\
        \midrule
        % \multirow{2}{*}{\rotatebox[origin=c]{90}{~} $\begin{dcases} \\  \end{dcases}$} &
        % %
        2D-TAN \cite{zhang2020learning} & 5.04 & 12.89 & 2.02 & 5.88\\
        %  %
        %  %
        VSLNet \cite{zhang2020span} & 5.45 & 10.74 & 3.12 & 6.63\\
        MS 2D-TAN \cite{zhang2021multi} & 7.05 & \textbf{14.15} & 4.75 & \textbf{9.16}\\
        Moment DETR \cite{lei2021qvhighlights} & 4.52 & 8.03 & 1.99 & 3.33\\
        Ours-variant (self-attention) & 6.53 & 11.02 & 4.05 & 7.59\\
        %  %
        %  %
        \midrule
        %  \multirow{6}{*}{\rotatebox[origin=c]{90}{~} $\begin{dcases} \\ \\ \\ \\ \\ \end{dcases}$}
        Ours-base & 7.69 & 11.51 & 4.83 & 7.8\\
        $\quad+$SW & 9.06 & 11.36 & 5.68 & 7.25\\
        $\quad+$VS & 7.38 & 10.66 & 4.31 & 6.84\\
        $\quad+$SW and VS   & 9.96 & 12.55 & 6.3 & 8.34\\
        $\quad+$SW, VS, and Contra & \textbf{10.79} & 13.19 & \textbf{6.74} & 8.85\\
        \bottomrule
    \end{tabular}
    \vspace{0.2cm}
    \caption{Performance of different methods on the val set.}
    \label{apptab:nlq_results}
\end{table}
All the experiments run on single NVIDIA Tesla V100 GPU. Unless otherwise specified, the default video features are Slowfast features. We implement previous state-of-the-art methods: VSLNet \cite{zhang2020span}, 2D TAN \cite{zhang2020learning}, Ms 2D-TAN \cite{zhang2021multi}, and Moment DETR \cite{lei2021qvhighlights} on the Ego4D NLQ dataset for comparison. 
We denote our multi-scale cross-modal transformer as Ours-base. We also denote Ours-base with variable-length sliding window sampling (SW), video splicing (VS), and video frame-level contrastive loss (Contra) as Ours-full.

The comparison results are shown in Table \ref{apptab:nlq_results}. Ours-full outperforms all state-of-the-art methods on R1@0.3 and R1@0.5. To verify the effectiveness of cross-attention mechanism, we replace the cross-attention mechanism on Ours-base with the self-attention mechanism as Moment DETR and denote it as Ours-variant (self-attention).
In Moment DETR, features of texts and video are concatenated along the sequence dimension before doing a self-attention operation. Compared with Ours-base, the performance of Ours-variant on all the metrics is degraded. It shows that using a cross-attention mechanism to explicitly interact video features with textual features can improve localization performance on Ego4D.

After adding a variable-length sliding window sampling strategy. The performance improved by 1.3\% on R1@0.3 without significant improvement in other metrics. 
When we use the video splicing strategy, the performance has not been improved, even a little worse. 
When we use the two data augmentation together, the performance on four metrics is boosted by 2.27\%, 1.04\%, 1.47\%, and 0.54\% compared with Ours-base. If we add video frame-level contrastive loss to this setting, the performance will reach the highest, and the four indicators are improved by 3.1\%, 1.68\%, 1.91\%, and 1.05\% compared to Ours-base respectively.

As can be seen from Table \ref{apptab:nlq_results}, 
Ours-full can achieve the best performance in Slowfast input. 
In addition, Ego4D provides video features extracted from two models: Slowfast and Omnivore. As shown in Table \ref{Ours-full}, Ours-full has a similar performance on the val set with these two different feature as input. However, the ensemble result has improved, that is, Ours-full-omnivore and Ours-full-slowfast are complementary. The ensemble strategy here is very simple. These two models output top5 results according to their prediction score (the format of the result is (start time, end time, score)), so there are 10 results. We sort these 10 results according to the score value and take the top5 as the final result.
For the final submission, we train Ours-full-slowfast and Ours-full-omnivore on the combination of train set and val set. The test set performance of our ensemble model achieves the best performance on R1@0.3 and R1@0.5 and competitive result on R5@0.3 and R5@0.5 as shown in Table \ref{test_results}.

\begin{table}[t]
    \centering
    \begin{tabular}{c c c c c}
        \toprule
        \multirow{2}{*}{Method} & \multicolumn{2}{c}{IoU=$0.3$ (\%)} & \multicolumn{2}{c}{IoU=$0.5$ (\%)}\\
        & R@1 & R@5 & R@1 & R@5\\
        \midrule
        % \multirow{2}{*}{\rotatebox[origin=c]{90}{~} $\begin{dcases} \\  \end{dcases}$} &
        % %
        Ours-full-slowfast &  10.79 & 13.19 & 6.74 & 8.85\\
        %  %
        %  %
        Ours-full-omnivore & 10.74 & 13.47 & 6.87 & 8.72\\
        \midrule
        Ensemble  & \textbf{11.33} & \textbf{14.77} & \textbf{7.05} & \textbf{8.98}\\
        \bottomrule
    \end{tabular}
    \vspace{0.2cm}
    \caption{Performance of our method with different video input feature on the val set.}
    \label{Ours-full}
\end{table}

\begin{table}[t]
    \centering
    \begin{tabular}{c c c c c}
        \toprule
        \multirow{2}{*}{Method} & \multicolumn{2}{c}{IoU=$0.3$ (\%)} & \multicolumn{2}{c}{IoU=$0.5$ (\%)}\\
        & R@1 & R@5 & R@1 & R@5\\
        \midrule
        % \multirow{2}{*}{\rotatebox[origin=c]{90}{~} $\begin{dcases} \\  \end{dcases}$} &
        % %
        Ensemble &  12.89 & 15.41 & 8.14 & 9.94\\
        \bottomrule
    \end{tabular}
    \vspace{0.2cm}
    \caption{Performance of our ensemble model on test set.}
    \label{test_results}
\end{table}
\textbf{Limitation.} Although our model can outperform other competitors on the R1 metric, the performance of our model in the R5 metric did not meet expectations.

\textbf{Acknowledgment.} This work is supported by National Key R\&D Program of China (No. 2020AAA0108800) and Fundamental Research Funds for the Central Universities (No. 226-2022-00051).

{\small
\bibliographystyle{ieee_fullname}
\bibliography{egbib}

\begin{thebibliography}{10}\itemsep=-1pt

\bibitem{feichtenhofer2019slowfast}
Christoph Feichtenhofer, Haoqi Fan, Jitendra Malik, and Kaiming He.
\newblock Slowfast networks for video recognition.
\newblock In {\em Proceedings of the IEEE/CVF international conference on
  computer vision}, pages 6202--6211, 2019.

\bibitem{girdhar2022omnivore}
Rohit Girdhar, Mannat Singh, Nikhila Ravi, Laurens van~der Maaten, Armand
  Joulin, and Ishan Misra.
\newblock Omnivore: A single model for many visual modalities.
\newblock {\em arXiv preprint arXiv:2201.08377}, 2022.

\bibitem{grauman2021ego4d}
Kristen Grauman, Andrew Westbury, Eugene Byrne, Zachary Chavis, Antonino
  Furnari, Rohit Girdhar, Jackson Hamburger, Hao Jiang, Miao Liu, Xingyu Liu,
  et~al.
\newblock Ego4d: Around the world in 3,000 hours of egocentric video.
\newblock {\em arXiv preprint arXiv:2110.07058}, 3, 2021.

\bibitem{lei2021qvhighlights}
Jie Lei, Tamara~L Berg, and Mohit Bansal.
\newblock Qvhighlights: Detecting moments and highlights in videos via natural
  language queries.
\newblock {\em arXiv preprint arXiv:2107.09609}, 2021.

\bibitem{radford2021learning}
Alec Radford, Jong~Wook Kim, Chris Hallacy, Aditya Ramesh, Gabriel Goh,
  Sandhini Agarwal, Girish Sastry, Amanda Askell, Pamela Mishkin, Jack Clark,
  et~al.
\newblock Learning transferable visual models from natural language
  supervision.
\newblock In {\em International Conference on Machine Learning}, pages
  8748--8763. PMLR, 2021.

\bibitem{vaswani2017attention}
Ashish Vaswani, Noam Shazeer, Niki Parmar, Jakob Uszkoreit, Llion Jones,
  Aidan~N Gomez, {\L}ukasz Kaiser, and Illia Polosukhin.
\newblock Attention is all you need.
\newblock {\em Advances in neural information processing systems}, 30, 2017.

\bibitem{wang2021t2vlad}
Xiaohan Wang, Linchao Zhu, and Yi Yang.
\newblock T2vlad: global-local sequence alignment for text-video retrieval.
\newblock In {\em Proceedings of the IEEE/CVF Conference on Computer Vision and
  Pattern Recognition}, pages 5079--5088, 2021.

\bibitem{yang2021multiple}
Yi Yang, Yueting Zhuang, and Yunhe Pan.
\newblock Multiple knowledge representation for big data artificial
  intelligence: framework, applications, and case studies.
\newblock {\em Frontiers of Information Technology \& Electronic Engineering},
  22(12):1551--1558, 2021.

\bibitem{zhang2021natural}
Hao Zhang, Aixin Sun, Wei Jing, Liangli Zhen, Joey~Tianyi Zhou, and Rick
  Siow~Mong Goh.
\newblock Natural language video localization: A revisit in span-based question
  answering framework.
\newblock {\em IEEE transactions on pattern analysis and machine intelligence},
  2021.

\bibitem{zhang2020span}
Hao Zhang, Aixin Sun, Wei Jing, and Joey~Tianyi Zhou.
\newblock Span-based localizing network for natural language video
  localization.
\newblock {\em arXiv preprint arXiv:2004.13931}, 2020.

\bibitem{zhang2021multi}
Songyang Zhang, Houwen Peng, Jianlong Fu, Yijuan Lu, and Jiebo Luo.
\newblock Multi-scale 2d temporal adjacency networks for moment localization
  with natural language.
\newblock {\em IEEE Transactions on Pattern Analysis and Machine Intelligence},
  2021.

\bibitem{zhang2020learning}
Songyang Zhang, Houwen Peng, Jianlong Fu, and Jiebo Luo.
\newblock Learning 2d temporal adjacent networks for moment localization with
  natural language.
\newblock In {\em Proceedings of the AAAI Conference on Artificial
  Intelligence}, volume~34, pages 12870--12877, 2020.

\bibitem{zhao2022centerclip}
Shuai Zhao, Linchao Zhu, Xiaohan Wang, and Yi Yang.
\newblock Centerclip: Token clustering for efficient text-video retrieval.
\newblock {\em arXiv preprint arXiv:2205.00823}, 2022.

\end{thebibliography}
}

\end{document}